\documentclass{article}
\usepackage{spconf,amsmath,epsfig}
\usepackage{caption}
\usepackage{graphicx}
\usepackage{float} 
\usepackage{subfigure}
\usepackage{subcaption}
\usepackage{tabularx}
\usepackage{multirow}
\usepackage{diagbox}
\usepackage{enumitem}
\usepackage{amsmath}
\usepackage{amsfonts}
\usepackage{overpic}
\usepackage{color}
\usepackage{xcolor}
\usepackage{pgfplots}
\usepackage{booktabs} 
\usepackage{url}

\usepackage{hyperref}

\usepackage[resetlabels]{multibib} 
\newcites{appendix}{Reference}

\let\OLDthebibliography\thebibliography
\renewcommand\thebibliography[1]{
  \OLDthebibliography{#1}
  \setlength{\parskip}{0pt}
  \setlength{\itemsep}{0pt plus 0.3ex}
}

\begin{document}\sloppy

\def\x{{\mathbf x}}
\def\L{{\cal L}}

\title{Training-Free Semantic Video Composition \\via Pre-trained Diffusion Model}
%
\name{
Jiaqi Guo,
Sitong Su,
Junchen Zhu,
Lianli Gao,
Jingkuan Song
}
\address{
University of Electronic Science and Technology of China (UESTC)\\
{\tt\small 
{jiaqiguo7@outlook.com}, 
{sitongsu9796@gmail.com}, 
{junchen.zhu@hotmail.com}
}
}

\maketitle

\begin{abstract}
The video composition task aims to integrate specified foregrounds and backgrounds from different videos into a harmonious composite. 
Current approaches, predominantly trained on videos with adjusted foreground color and lighting, struggle to address deep semantic disparities beyond superficial adjustments, such as domain gaps.
Therefore, we propose a training-free pipeline employing a pre-trained diffusion model imbued with semantic prior knowledge, which can process composite videos with broader semantic disparities.
Specifically, we process the video frames in a cascading manner and handle each frame in two processes with the diffusion model. In the inversion process, we propose Balanced Partial Inversion to obtain generation initial points that balance reversibility and modifiability. Then, in the generation process, we further propose Inter-Frame Augmented attention to augment foreground continuity across frames.
Experimental results reveal that our pipeline successfully ensures the visual harmony and inter-frame coherence of the outputs, demonstrating efficacy in managing broader semantic disparities. 

\end{abstract}
\begin{keywords}
video composition, training-free, semantic disparities.
\end{keywords}
\section{Introduction}
\label{sec:intro}

\begin{figure}[t]
    \centering
    \includegraphics[width=1\linewidth]{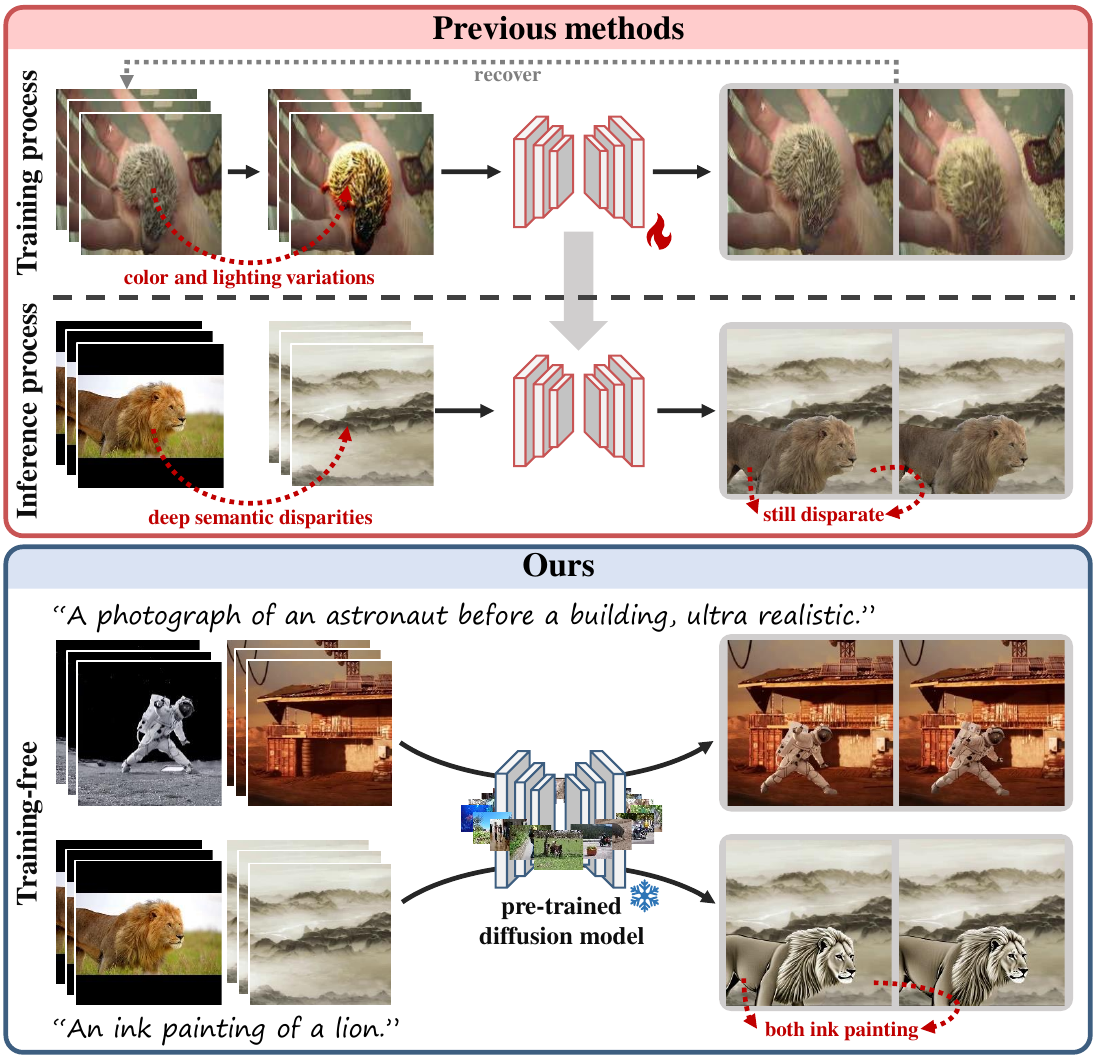}
    \caption{Comparison of our methods with previous methods. \textbf{Above:} Training and inference process of previous methods\cite{CO22022deep}. They perform poorly facing deep semantic disparities. \textbf{Below:} Our training-free pipeline. We achieve satisfactory results in both color and lighting adjustments and deep semantic transformation. More cases can be found in \url{https://anonymous.4open.science/r/paper130}.}
    \label{fig1:first}   
\end{figure}

Video composition investigates seamlessly blending multiple specified objects into a visually harmonious video, which has extensive applications in social media, artistic creation, and film production. It can be categorized into two types: text-guided composition and reference-guided composition. 
Text-guided composition\cite{dreamix2023dreamix} is provided with an image or video of the scenario and sentences of specific objects, thereby generating matching videos on this scenario. 
In contrast, reference-guided composition\cite{wang2017copy, wang2019illumination} presents greater challenges since it provides reference videos of specific objects and requires the output to restore detailed characteristics. 


Existing methodologies of reference-guided video composition primarily concentrate on modifying color and lighting of the specified foreground to achieve visual concordance with the background, which is also referred to as video harmonization\cite{CO22022deep, huang2019temporally, Harmonizer2022harmonizer}. These approaches often train a network on extensive samples which are typically constructed by artificially introducing disharmony in the foreground appearance of ground-truth samples, as shown in the first row of Fig. \ref{fig1:first}. In this strategy, the networks are adept at adjusting the pixel colors of the foreground and recovering the training samples. 
However, a significant limitation arises when encountering more complex semantic disparities between the specified foreground and background. An example of such a disparity is given in Fig.\ref{fig1:first}: when the foreground comprises a real-world lion and the background originates from an ink painting, these methods struggle to effectively integrate the two elements.

Consequently, We argue that reference-guided video composition should move beyond mere color and lighting adaptation, extending its focus to encompass deeper semantic disparities. To achieve this, we propose a practical pipeline for video composition by introducing a large-scale pre-trained diffusion model for its extensive semantic prior knowledge. Our pipeline is illustrated in Fig.\ref{fig2:pipeline}. 

Overall, we sequentially process the video frame-by-frame and employ the latent diffusion model\cite{sd2022high} as the backbone to handle each frame through two processes called inversion and generation. More precisely, for each composite frame, we first utilize Balance Partial Inversion (BPI) during the inversion process to produce an initial point for generation. This initial point is a specific latent feature that can be modified by conditions while maximizing the preservation of the characteristics of the current frame. Then, starting from this initial point, the generation process is performed to produce the result, as shown in the yellow box in Fig.\ref{fig2:pipeline}. During the generation process, we utilize Inter-Frame Augmented attention (IFA) to establish inter-frame linkages and augment the continuity of the foregrounds in the generated frames.

To summarize, our major contributions are as follows:
\begin{itemize}
\setlength{\topsep}{1pt}
\setlength{\itemsep}{1pt}
\setlength{\parsep}{1pt}
\setlength{\parskip}{1pt}
    \item We propose a training-free pipeline for reference-guided video composition to handle various semantic disparities beyond simple color and lighting adjustments.
    \item We present the Balanced Partial Inversion, which can provide appropriate generation initial points that accommodate both reversibility and modifiability for diffusion-based composition methods.
    \item Extensive experiments prove that our pipeline can process not only superficial visual differences but also deep semantic disparities in composite videos.
\end{itemize}

\section{related work}

Traditional reference-guided \textit{video composition} is a hot topic in the field of video processing. Researchers tend to focus on using mathematical methods\cite{motion2013motion, wang2017copy, wang2019illumination} such as Possion blending or mean-value cloning to improve the quality of composites. 
The popularity of neural networks has allowed training with large datasets to become the mainstream approach. 
Huang et al.\cite{huang2019temporally} proposed to apply affine transforms to the foregrounds of composite images and acquire a series of images containing the same foreground as videos for training. Lu et al.\cite{CO22022deep} proposed the first public dataset by collecting a large number of videos and adjusting their foregrounds to simulate composite videos. However, these data limit the capabilities of models primarily to color and lighting adjusting rather than semantic adaptation. We aim to enrich the capabilities by leveraging the rich prior knowledge embedded in large-scale pre-trained models.

Current reference-guided \textit{image composition} approaches generally adopt two paradigms: harmonization\cite{CDTnet2022high, PCT2023pct, Harmonizer2022harmonizer, LEMART2023lemart} and blending\cite{PaintbyExample2023paint, PHD2023paste}. The former, however, struggles with complex semantic gaps\cite{PIHnet2023semi, SSH2021ssh}, similar to the challenges faced in video harmonization. On the other hand, while the blending paradigm offers robust visual and semantic consistency\cite{anydoor2023anydoor, TFICON2023tf, PHD2023paste}, its application to video frequently results in severe inter-frame deformation and flickering. 
Our method aims to navigate these challenges, effectively balancing the complex semantic differences handling and inter-frame deformation issues.

\section{method}

\begin{figure*}[t]
    \centering
    \includegraphics[width=1\linewidth]{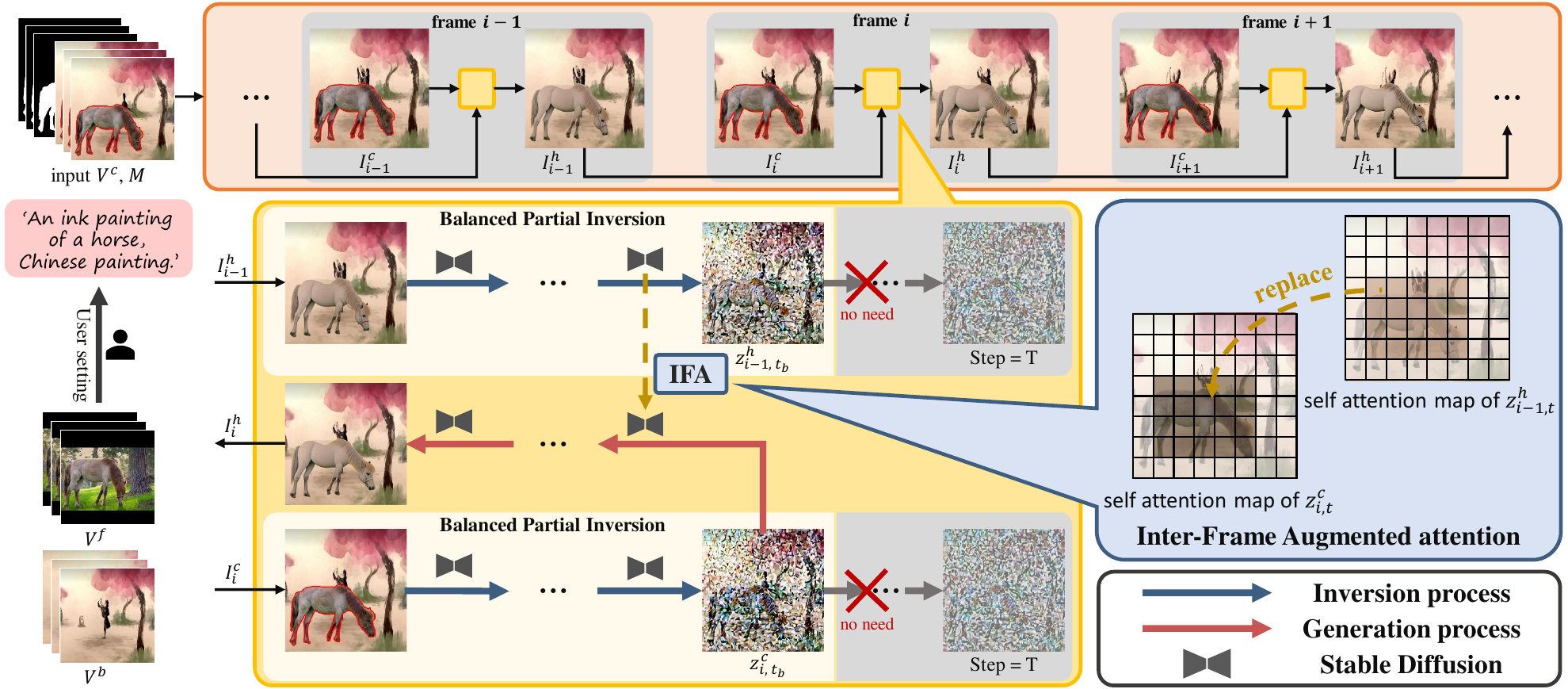}
    \caption{Our proposed training-free pipeline. We process the composite video $V^c$ frame-by-frame in a cascading manner, as shown in the \textcolor{orange}{orange box} at the top of the figure. The \textcolor{yellow}{yellow box} illustrates our process for each frame. Specifically, we employ the Stable Diffusion\cite{sd2022high} to process frame $i$ in two processes: \textit{inversion} and \textit{generation}. During the \textit{inversion} process, we invert the $I^c_i$ in $t_b$ steps to obtain an initial point $z^c_{i, t_b}$ using \textbf{Balanced Partial Inversion (BPI)}. Then, we start the generation process from this initial point. During the \textit{generation} process, the processed previous frame $I^h_{i-1}$ affects the current frame through the \textbf{Inter-Frame Augmented attention (IFA)} to associate frames with each other, which is shown in the \textcolor{cyan}{blue box}. }
    \label{fig2:pipeline}
\end{figure*}

\begin{figure}[t]
    \centering
    \includegraphics[width=1\linewidth]{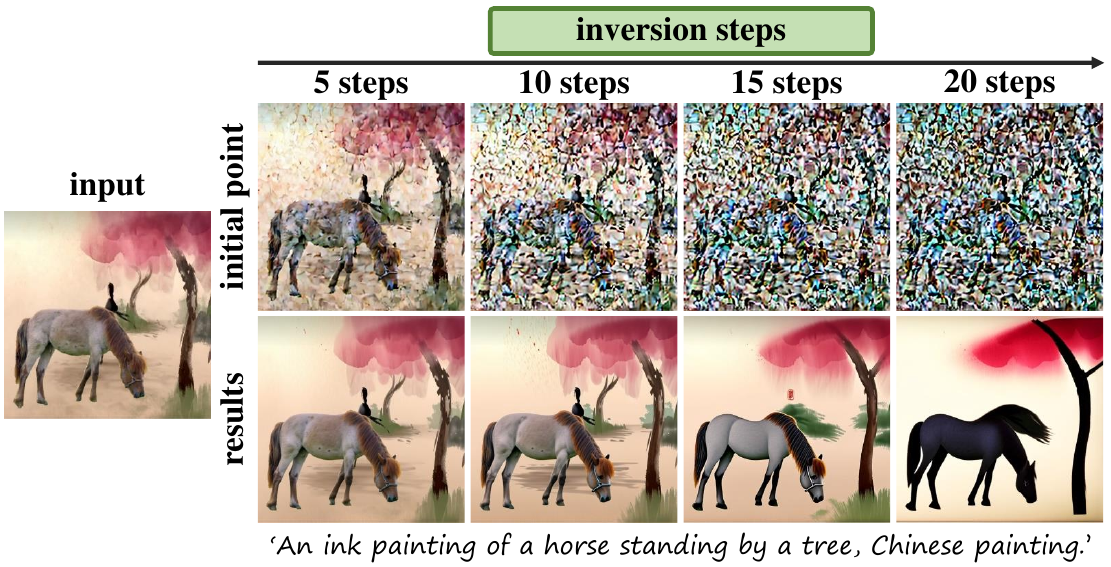}
     \caption{Image reconstruction using the latent of different inversion steps as the initial point. The complete inversion process takes $T=20$ steps. The reconstructed image generated from the initial point with fewer inversion steps will retain more characteristics of the input.}
    \label{fig:method}
\end{figure}

\begin{figure*}[t]
    \centering
    \includegraphics[width=1\linewidth]{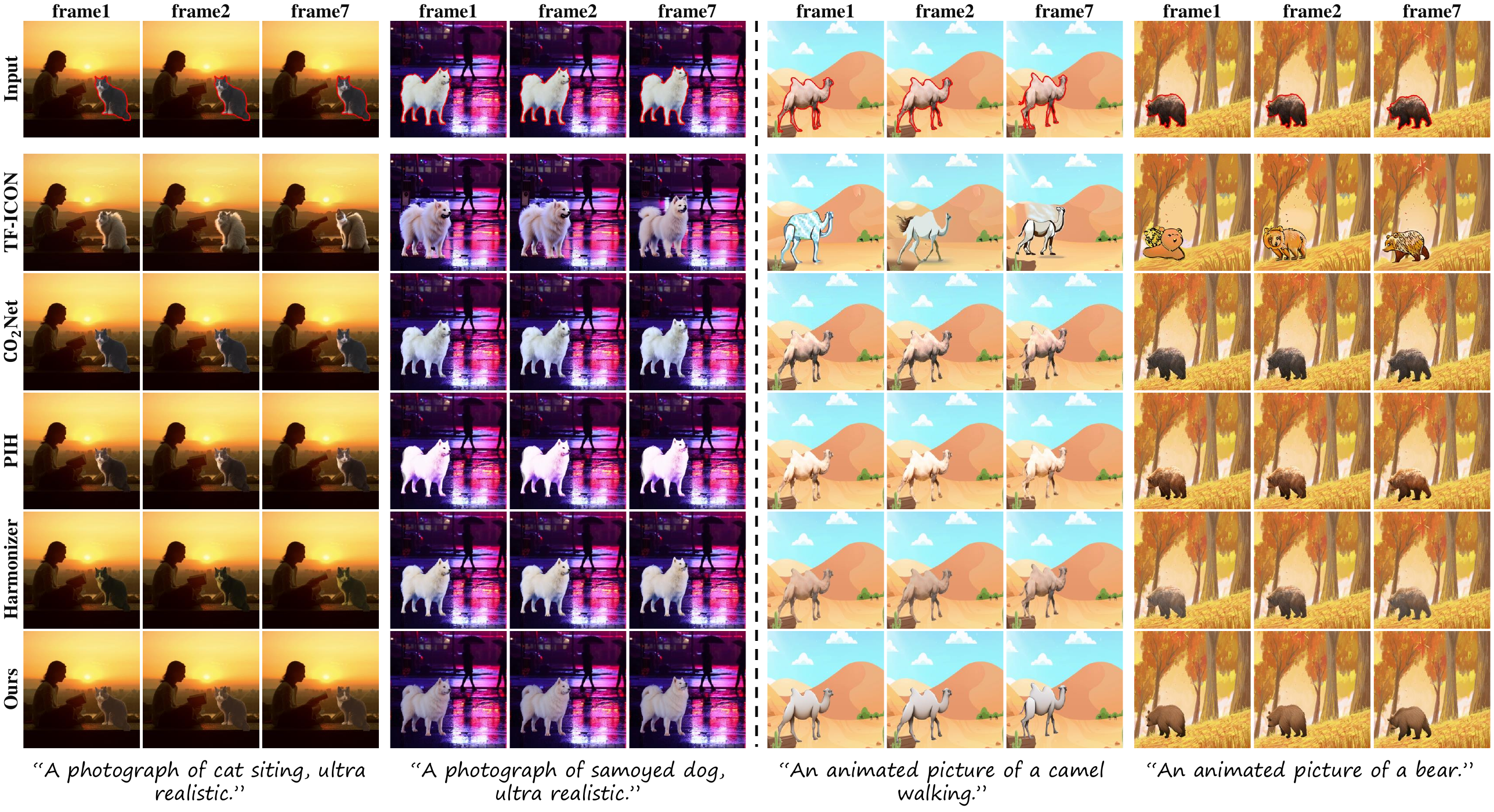}
    \caption{Qualitative comparison with methods of image harmonization (PIH and Harmonizer), video harmonization (CO$_2$Net), and image blending (TF-ICON). There are four examples in total. The two on the left are composites needing color and lighting adjustments and the two on the right are composites with deep semantic disparities. The text conditions are listed at the bottom of the figure (only needed in TF-ICON and Ours).}
    \label{fig3:comparison}
\end{figure*}

\subsection{Method Overview}

\noindent\textbf{Problem Definition.} 
Given the reference foreground video $V^f$, after adjusting its scale and position according to user settings, it is pasted to the background video $V^b$ to obtain the preliminary composite video $V^c = \{I^c_i\}^n_{i=1}$. The mask corresponding to the specified foreground in $V^c$ is defined as $M = \{M_i\}^n_{i=1}$. Our training-free pipeline symbolized by $\mathcal{H}$ aims to transform $V^c$ into a visually harmonious and semantically consistent video $V^h = \{I^h_i\}^n_{i=1}$ with the textual description $\mathcal{P}$ of the desired semantics. The whole process can be represented by Equ.\ref{equ1}.
\begin{equation}
    \begin{split}
        V^c &= V^f \cdot M + V^b \cdot  (1 - M), \\
        V^h &= \mathcal{H}(V^c, M, \mathcal{P}).
    \end{split}
\label{equ1}
\end{equation}

\noindent\textbf{Framework Overview.} 
Our pipeline processes the preliminary composite video $V^c$ on a frame-by-frame basis using a pre-trained text-to-image latent diffusion model, also known as Stable Diffusion\cite{sd2022high}. Each frame, except the first, is assisted by the generated result of the previous frame during its own generation phase. The processing of each frame consists of two main processes: inversion and generation. In addition, We propose two strategies, Balanced Partial Inversion (BPI) and Inter-Frame Augmented attention (IFA), to respectively refine these two processes in order to effectively produce the desired outcomes, as depicted in Fig. \ref{fig2:pipeline}. 

In detail, for each frame $I^c_i$, BPI is applied during the inversion process to derive an initial point that enables semantic adjustments while preserving the detailed information of the input frame. Subsequently, the generation process commences from this initial point. IFA provides the current frame $I^c_i$ with information from the processed previous frame $I^h_{i-1}$, achieved by replacing the foreground segment of the self-attention maps at each network layer during the generation process. Next, we describe these two processes separately.

\subsection{Inversion Process} 
Diffusion-based models generate image by sequentially removing noise from an initial point, usually a Gaussian noise, across $T$ steps. Some image-editing related tasks require finding a non-randomized initial point that produces the given input image, a process known as inversion\cite{p2phertz2022prompt}. In the context of reference-guided video composition, maintaining the characteristics of the reference videos is essential, which necessitates effective inversion strategies to secure qualified initial points. Several researchers modify the solvers\cite{diffedit2022diffedit, diffusionclip2022diffusionclip} to optimize the inversion process. In contrast to these complicated and mathematical methods, we find that simply using the result of intermediate timestep of the inversion process as the generation initial point can well preserve the characteristics of the input image. 

\noindent\textbf{Balanced Partial Inversion.} 
As depicted in Fig.\ref{fig:method}, we conduct image reconstruction experiments with different inversion steps as initial points. 
A lower number of inversion steps generates an initial point that retains more information about the input image, resulting in a reconstructed result that closely resembles the original input. However, these results are resistant to be modified by the given conditions. For example, as shown in Fig.\ref{fig:method}, when the inversion step is 5, the text condition has minimal impact.
Conversely, a greater number of inversion steps makes the result more susceptible to be altered by text or other conditions, thus losing the restoration of details. 

In view of this, when processing each frame $I^c_i$, we use the partial inversion result $z^c_{i,t_b}$ as the initial point for generation. $t_b$ represents the inversion step that generates an initial point which achieving a balance between preserving details of the input frame and allowing for alteration.
The exact number of inversion steps $t_b$ ($t_b \in (0,T) $) is determined by the degree of the semantic disparities in the composite video.

\subsection{Generation Process}  
Commencing with $z^c_{i,t_b}$ as the initial point and utilizing the desired text descriptions $\mathcal{P}$ as a conditional guide, the generation process unfolds by progressively removing noise from the initial point across $t_b$ steps. Given that these textual descriptions primarily dictate the overarching semantic style without providing fine-grained control, we adopt a technique\cite{TFICON2023tf} which involves calculating the cross-attention maps between foregrounds and backgrounds to further guide the generation of each frame. 

\noindent\textbf{Inter-Frame Augmented Attention.} 
While individual frame processing effectively preserves specific information within each frame, this approach can lead to the loss of relational information between frames. This may result in incoherence in the processed videos, such as deformations in the foreground between neighboring frames. Therefore, we propose Inter-Frame Augmented attention to strengthen the correlation between frames. 

As shown in Fig.\ref{fig2:pipeline}, during the processing of the current frame $I^c_i$, the previously processed frame $I^h_{i-1}$ undergoes simultaneous inversion to obtain $z^h_{i-1,t_b}$. The connections are then established by replacing the foreground region of the self-attention map.

Specifically, at timestep $t$ of the generation process, the UNet conducts a denoising operation on the latent feature $z^c_{i,t}$. At each layer $l$, we respectively compute the self-attention maps for $z^c_{i,t}$ and $z^h_{i-1,t}$, denoted as $A_{i,t,l}$ and $A_{i-1,t,l}$. The mask $M_i$ is then scaled to align with the dimensions of the features in layer $l$, and is used to identify the foreground location in $A_{i,t,l}$ and form a new mask denoted as $M^a_{i,l,t}$. We then use $M^a_{i,l,t}$ to make the replacement by:
\begin{equation}
    A'_{i, l, t} = A_{i, l, t} \cdot (1 - M^a_{i, l,t}) + A_{i-1, l, t} \cdot M^a_{i, l,t}.
\label{equ2}
\end{equation}
Note that the IFA does not work in all timesteps of the generation process. In order to balance the attention to the previous frame and the attention to the current generating frame, we set a threshold $\tau$ to determine the operating range of IFA ($t \in [\tau, t_b]$).

\noindent\textbf{Background Replace.} While BPI enables the generated results to retain the characteristics of the input frames well, the injection of textual prompts and cross-attention maps often results in deviations from the reference background. Therefore, to further preserve the background, we directly use the mask $M_i$ to replace it in the final generated result with the reference background.

\section{experiments}

In this section, we first outline the datasets and metrics utilized for experimental validation. Then we qualitatively and quantitatively show the comparative analysis of our method against previous approaches. 
Lastly, through ablation studies, we validate the efficacy of the various components integral to our method.

\subsection{Settings}
\label{settings}

\noindent\textbf{Test Dataset.} Existing composite video datasets\cite{CO22022deep} are constructed by manually adjusting the color and lighting of the foregrounds of ground-truth videos, which can not evaluate the abilities to handle composition with diverse semantic differences. Therefore, we have collected 15 composite videos with shallow semantic disparities and 12 composite videos with deep semantic disparities from DAVIS2017\cite{davis20172017} and from the web as a test dataset. Shallow semantic disparities imply that the foregrounds and backgrounds originate from different videos but the same domain (both realism), necessitating only color and lighting adjustments to achieve visual harmony. In contrast, deep semantic disparities indicate that the foregrounds and backgrounds are derived not only from different videos but also from distinct domains, including ink paintings, animations, and realism. Each sample in this dataset has 10 frames and consists of a background video, a foreground video with its corresponding mask, and a text prompt.
\\

\noindent\textbf{Metric.} With no ground-truth available for our task to measure the quality of the outputs, we compute the metrics directly from the outputs from two perspectives: (1) the coherence of the video frames and (2) the semantic difference between foregrounds and backgrounds. 
For the inter-frame coherence, we follow the practice of previous works\cite{CO22022deep, huang2019temporally} and use Temporal Loss as the metric. 
Lower values indicate better coherence between video frames.
For the semantic difference, borrowing from metrics of style transfer, we extract the features of the outputs and the reference backgrounds with VGG-19\cite{VGG2014very}, and use the difference between their Gram matrices as the metric. Lower values indicate smaller semantic differences between foregrounds and backgrounds.
\\

\subsection{Qualitative Comparison with SOTA Methods}

To intuitively demonstrate the effectiveness of different approaches, we show the visualization results compared with the previous methods in Fig.\ref{fig3:comparison}. Based on the target task, existing baselines can be categorized into three groups: image harmonization, image blending, and video harmonization. We selected methods from all the three groups for comparison, including PIH\cite{PIHnet2023semi}, Harmonizer\cite{Harmonizer2022harmonizer}, TF-ICON\cite{TFICON2023tf}, and CO$_2$Net\cite{CO22022deep}. Input denotes directly extracting the foreground and pasting it onto the background video without any further processing. 

PIH\cite{PIHnet2023semi}, Harmonizer\cite{Harmonizer2022harmonizer} and CO$_2$Net\cite{CO22022deep} are methods designed for the harmonization task. Since they are trained to reconstruct foreground colors of manually color-tuned videos, they can not tone well when the foregrounds come from other images that are irrelevant to the backgrounds. They perform even more ineptly when faced with composite videos that require deeper semantic adjustments. For instance, in the camel and bear example in Fig.\ref{fig3:comparison}, they can not seamlessly integrate realistic foregrounds into animated backgrounds. TF-ICON\cite{TFICON2023tf} introduces textual information in the same way as our approach to handily handle composition with different semantic disparities. However, it is not good at preserving the appearance and characteristics of the reference foregrounds and fails to achieve inter-frame coherence. In contrast, our approach better adjusts the color and semantics of the foregrounds while preserving the appearance well. 

\begin{table}[t]
\caption{Quantitative comparisons for video composition. We calculate the inter-frame coherence (Temporal Loss, TL) and the semantic differences between foregrounds and backgrounds (Semantic Loss, SL) of the outputs. The optimal and suboptimal results are bolded and underlined, respectively.} \label{tab:cap}
\centering
{\small 
    \begin{tabular}{ccccccccc}
    \toprule[1pt]
    \begin{tabular}[c]{@{}c@{}}Metric\\\tiny$\times10^3$\end{tabular} & Tf-ICON    & CO$_2$Net    & PIH    & Harmonizer    & Ours \\ \hline
    \rule{0pt}{3ex}
    TL$\downarrow$ & 127.30  & \underline{8.33}  & 18.62  & 15.17  & \textbf{7.51} \\ 
    \rule{0pt}{3ex}
    SL$\downarrow$ & \textbf{57.34}  & 79.59  & 94.69  & 82.10  & \underline{73.91} \\ 
    \bottomrule[1pt]
    \end{tabular}
    \label{tab1:quantitative}
} 
\end{table}

\subsection{Quantitative Comparison with SOTA Methods}
 
The quantitative comparison results are listed in Table \ref{tab1:quantitative}. TF-ICON\cite{TFICON2023tf} obtained the best score on the semantic differences measure but the worst score on the inter-frame coherence, similar to the conclusion in Fig.\ref{fig3:comparison}: while it adeptly manages the semantic disparities between foreground and background, it can struggle to maintain foreground coherence across adjacent frames. PIH\cite{PIHnet2023semi} and Harmonizer\cite{Harmonizer2022harmonizer}, designed for image harmonization task, exhibit discrepancies in harmonized color across different frames when applied to video, resulting in a slightly worse inter-frame coherence compared to video harmonization method CO$_2$Net\cite{CO22022deep}. However, all the three methods demonstrate underperformance in semantic difference measures. In comparison, our method strikes a better balance between inter-frame coherence and semantic consistency, enabling more desirable generation results.

\subsection{Ablation Study}

To demonstrate the effectiveness of our key design choices, we abated our pipeline in five stages:
(1) \textbf{Baseline}: the frames are generated from the Gaussian noises, which are obtained by completely inverting the preliminary composite frames. (2) \textbf{+BPI}: Balanced Partial Inversion is applied to obtain the initial point of generation. (3) \textbf{+cross attention}: The cross-attention maps are injected during the generation process. (4) \textbf{+IFA}: Inter-Frame Augmented attention is further applied during the generation process. (5) \textbf{+bg replace}: The background in the final result is replaced to make it consistent with the reference background.

\begin{figure}[t]
    \centering
    \includegraphics[width=1\linewidth]{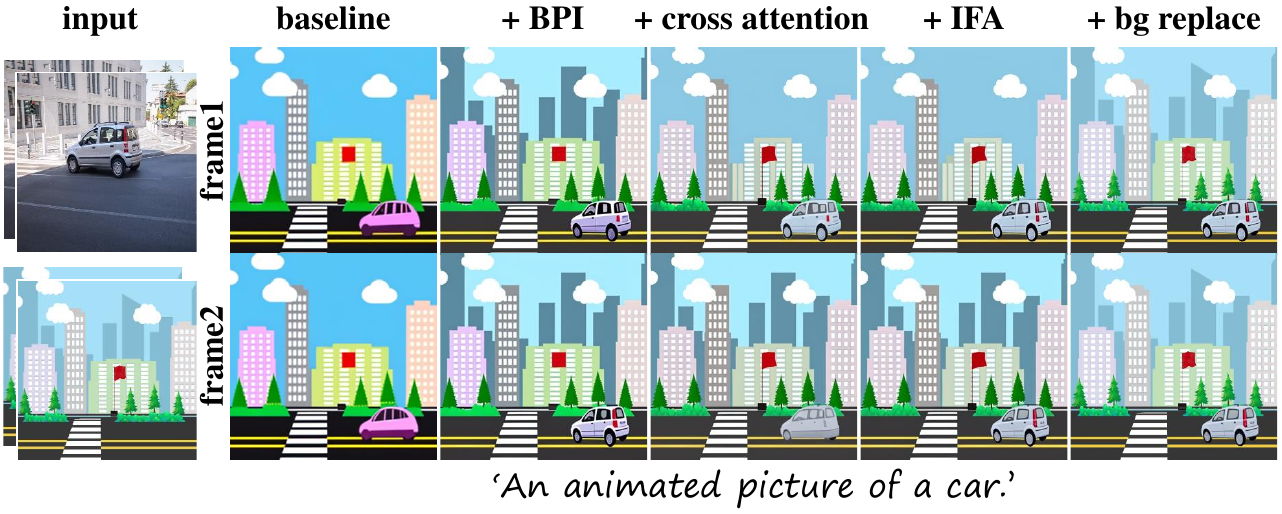}
     \caption{Ablation study of different variants of our framework. \textbf{BPI}: Balanced Partial Inversion. \textbf{IFA}: Inter-Frame Augmented attention. \textbf{bg}: background.}
    \label{fig5:ablation}
\end{figure}

Figure \ref{fig5:ablation} represents the visualization of each stage. Compared to the baseline, BPI is beneficial in maintaining a balance between narrowing the semantic disparities and preserving the shape of the reference foreground and background. On the other hand, IFA enables better foreground continuity of the processed results of neighboring frames. Overall, our full pipeline (last column in Fig.\ref{fig5:ablation}) preserves the characteristics of the reference videos better, and results in more visually harmonious and inter-frame coherent outputs.

\section{conclusion}

In this work, we propose a training-free pipeline to overcome the limitations in video composition task and deliver visually pleasing outcomes in compositing videos with various semantic disparities. We enhance the workflow of diffusion-based composition models with Balanced Partial Inversion and Inter-Frame Augmented attention. Our pipeline outperforms other methods on the test dataset. 
As a future work, we would like to explore the potential to generalize to multi-object video composition and further extend the diversity of video composition.



\bibliographystyle{IEEEbib}
{\small 
\bibliography{video_harmonization_simple}
} 

\clearpage
\appendix
\setcounter{figure}{0}

\noindent\makebox[\columnwidth]{\textbf{\large APPENDIX}}

\section{implementation details}
Our framework is built upon the Stable Diffusion v2-1\citeappendix{appsd2022high} architecture. We standardize all composite frames to a resolution of $512 \times 512$. In the inversion process, the exceptional prompt technique\citeappendix{appTFICON2023tf} is employed to mitigate the influence of textual descriptions. For composites where both foreground and background elements are realistic, a Look-Up Table (LUT)\citeappendix{appCO22022deep} is utilized to enhance inter-frame continuity at the pixel level in the final outputs. The optical flow is extracted using FlowNet2\citeappendix{appflownet22017} when calculating the Temporal Loss (TL) metric. For testing, some data are sourced from the web, with the Segment Anything Model\citeappendix{appsegany2023segany} employed to isolate the requisite foreground elements.

\section{additional analysis}
In this section, we delve deeper into the parametric variations of Balanced Partial Inversion (BPI) and Inter-Frame Augmented Attention (IFA) respectively through expanded experimental analysis.

\begin{figure}[htbp]
    \centering
    \begin{overpic}[width=1\linewidth]{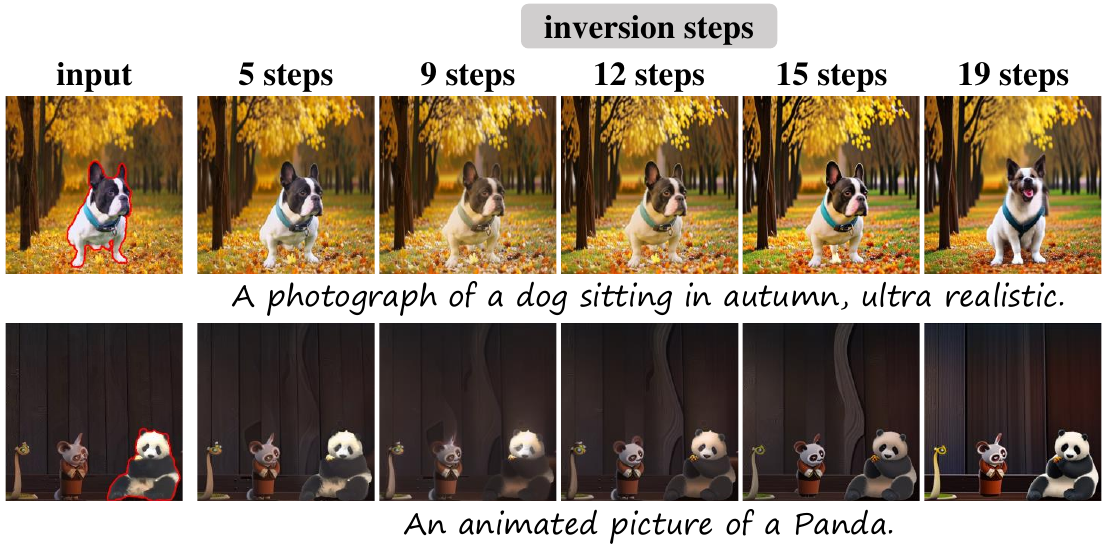}
        \put(34,25){\color{red} \linethickness{1pt} \framebox(17, 16){}}  
        \put(67,4.5){\color{red} \linethickness{1pt} \framebox(17, 16){}}
    \end{overpic}
    \caption{Compositing results using the latent of different inversion steps as the initial point. The complete inversion process takes 20 steps. Two examples are shown, each showing one frame. The best results are marked with red boxes.}
    \label{fig:BPI}
\end{figure}

\noindent\textbf{Inversion steps of BPI.}  As depicted in Fig.\ref{fig:BPI}, we analyze the impact of varying inversion steps as the initial point for video composition in scenarios with both shallow and deep semantic disparities. Consistent with the findings presented in Fig.3 of the main text, fewer inversion steps better preserve original characteristics but offer less modifiability. In cases with shallow semantic disparities (e.g., the first example in Fig.\ref{fig:BPI}), few steps, such as 9, suffice due to the limited need for editing only foreground color and lighting. However, deeper semantic disparities require more inversion steps for altering profound foreground features (e.g., in the second example the best is 15 steps). Importantly, an excessive number of steps might make the model can not accurately reconstruct the original frame, leading to deviations. For instance, in the second example with 19 inversion steps, even though the foreground fits the description of "animated" very well, the generated result is overly bright, which will result in a lack of harmony when the original background is replaced.

\begin{figure}[htbp]
    \centering
    \begin{overpic}[width=1\linewidth]{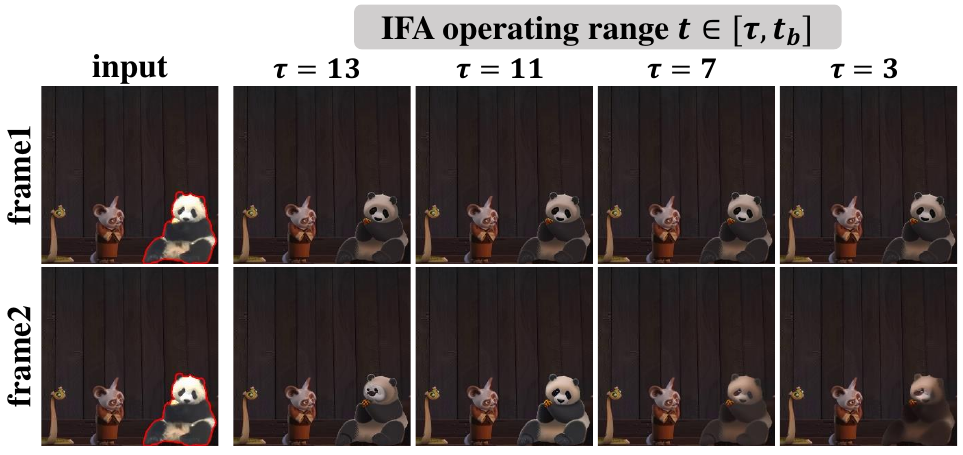}
        \put(43,1){\color{red} \linethickness{1pt} \framebox(19, 40.5){}}  
        \put(33,12){\color{yellow} \linethickness{1pt} \vector(1,-1){4}}
        \put(90,12){\color{yellow} \linethickness{1pt} \vector(1,-1){4}}
    \end{overpic}
    \caption{Compositing results with different operating range of IFA. The complete generation process takes 20 steps. In this case, $t_b=15$. The best results are marked with a red box.}
    \label{fig:IFA}
\end{figure}

\noindent\textbf{Operating range of IFA.} Fig.\ref{fig:IFA} illustrates the outcomes for different IFA operating ranges. A narrow operational range fails to maintain consistency in the foreground across adjacent frames. For instance, at $\tau=13$, significant deformation is observed in the panda's face in the frame 2 (as indicated by the yellow arrow). 
Conversely, an overly broad range leads to excessive focus on the previous frame, resulting in blurring or artifacts. For example, at $\tau=3$, the foreground of the frame 2 appears blurred.

\section{more cases}
\begin{figure*}[htbp]
    \centering
    \includegraphics[width=1\linewidth]{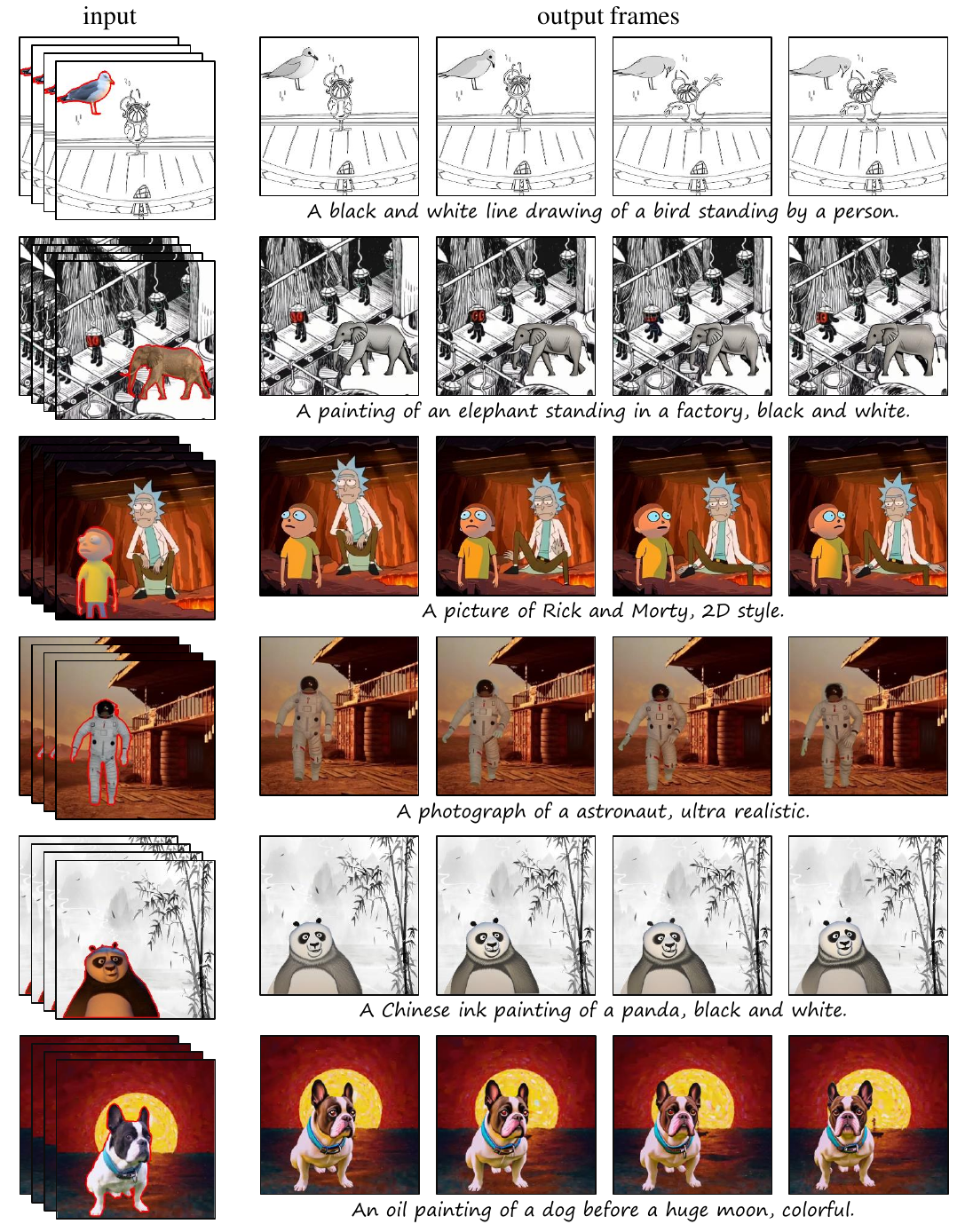}
     \caption{Composite videos with broader semantic disparities.}
    \label{fig:more cases}
\end{figure*}

Fig.\ref{fig:more cases} presents additional examples of video composition encompassing a wider range of semantic disparities, including composites between realism, line drawing, animation, ink painting, oil painting, as well as 2D and 3D styles. 

\bibliographystyleappendix{IEEEbib}
\bibliographyappendix{appendix}

\end{document}